\documentclass[conference]{IEEEtran}  

\usepackage{multirow}
\usepackage[pdftex]{graphicx}
\graphicspath{{./figures/}}
\usepackage{cite}
\usepackage{mdwmath} 
\usepackage{mdwtab}
\usepackage{eqparbox}
\usepackage{mathtools}
\usepackage[utf8]{inputenc}
\usepackage{multicol}
\usepackage[]{subcaption}  
\usepackage{amsmath}  
\usepackage{caption}

\usepackage{dsfont}
\usepackage{bm}
\usepackage{physics}  
\usepackage{amssymb}  
\usepackage{csvsimple}  
\usepackage{algorithm}
\usepackage{algpseudocode}
\usepackage{listings} 
\usepackage{amsfonts}

\usepackage{amsthm} 
\usepackage{gensymb} 
\usepackage{mathtools}
\usepackage{hyperref} 
\hypersetup{
  colorlinks   = true,    
  urlcolor     = blue,    
  linkcolor    = blue,    
  citecolor    = red      
}

\newtheorem{definition}{Definition}

\DeclareMathOperator{\E}{\mathbb{E}}

\newcommand{\argmin}{\arg\,\min}
\renewcommand{\pb}[1]{\left({#1}\right)} 

\newcommand{\sqb}[1]{\left[{#1}\right]} 
\newcommand{\nn}{\nonumber} 
\renewcommand{\norm}[1]{\left\lVert{#1}\right\rVert}

\allowdisplaybreaks
\begin{document}

\title{MrSARP: A Hierarchical Deep Generative Prior for SAR Image Super-resolution}
\author{Tushar Agarwal, Nithin Sugavanam
        and~Emre Ertin

\thanks{T. Agarwal, Nithin Sugavanam, and E. Ertin are with the Department
of Electrical and Computer Engineering, The Ohio State University, Columbus,
OH, 43210 USA}
\thanks{Corresponding Author: T. Agarwal (agarwal.270@buckeyemail.osu.edu)}
\thanks{This work has been submitted to the IEEE for possible publication. Copyright may be transferred without notice, after which this version may no longer be accessible.}
}

\maketitle 


\begin{abstract}
	Generative models learned from training using deep learning methods can be used as priors in inverse under-determined inverse problems, including  imaging from sparse set of measurements.  In this paper, we present  a  novel hierarchical deep-generative model MrSARP for SAR imagery that can synthesize  SAR images of a target at different resolutions jointly. MrSARP is trained in conjuction with a critic that scores multi resolution images jointly to decide if they are realistic images of a target at diffferent resolutions. We  show how this deep generative model can be used  to retrieve the high spatial resolution image  from low resolution images of the same target. The cost function of the generator is modified to improve its capability to retrieve the input parameters for a given set of resolution images. We evaluate the model's
	performance using the three standard error metrics used for evaluating
	super-resolution performance on simulated data and compare it to upsampling and sparsity based image sharpening approaches.
\end{abstract}

\begin{IEEEkeywords}
Deep Learning, Super-Resolution, Compressive sensing. 
\end{IEEEkeywords}
\IEEEpeerreviewmaketitle

\section{Introduction}\label{intro}
Synthetic aperture radar (SAR) imagery captures the physical aspects of the
target differently compared to the Optical imagery because of multi-path
reflections, specular nature of reflectors, and Imaging geometry effects leading
to overlay, shadowing. 
In this work, we present a generative model
that captures SAR phenomenology at different resolutions exhibited at SAR magnitude imagery. Specifically, we propose a specialized hierarchical architecture of the generative model,
called MrSARP, that jointly models the data manifold of multiple-resolutions of
a SAR image. We show that such a generative model acts as the projection operator to a lower dimensional manifold and can be directly used for super-resolving magnitude SAR images from low resolution magnitude images. The super-resolution performance of MrSARP is evaluated by comparing it with LASSO (or $L_1$ recovery) and Nearest Neighbor Upsampling using empirical data. 

Next, we describe important notation used throughout the text.
The lower-case character $x$ represents a fixed sample value of a random vector
(rv) $X$ (the corresponding upper-case character), $M:N$ is the enumeration of
natural numbers from $M$ to $N$, $g(X_{1:N};w)$ denotes function $g$ of rv's
$X_{1:N}$ with deterministic parameters $w$ and an index-based $x[v]$ represents
the $v^{th}$ value from the ordered set $\{x[i]: i \in \{1:N\}\}$. Formal definitions of terms Generative
Modeling and Compressed Sensing, as used in the context of this paper, are stated below.

\begin{definition}[Generative Modeling]\label{def:GM}
	Generative Modeling aims to estimate the distribution $p(X)$
	using a parameterized distribution family $q(X;w)$ and a set $\mathcal{D}$
	of samples from the distribution $p(X)$. The stochasticity in
	sampling is typically achieved by sampling a low-dimensional latent rv $Z$,
	internal to $q(X;w)$, from a simple distribution (e.g. standard gaussian).
	Therefore, the goal is to generate $N$ realistic samples $\{\hat{X}[v]\}_{v=1}^N$ from
	$q(X;w)$ as if they were from $p(X)$.
\end{definition}

\begin{definition}[Compressed Sensing]\label{def:CS}
    Given measurements $y \in \mathbb{C}^d$ obtained using a known measurement
    (or forward) operator $F$ of an underlying signal $x \in \mathbb{C}^D$, Compressed
    Sensing seeks to recover this underlying signal under the model 
    \begin{align}
        y = F(x) + \eta, \qquad x \in \Omega,\label{eq:CS_prom1}
    \end{align}
    where $\eta$ is the measurement noise and $\Omega$ represents the constraint
    set that is non-convex. It is an under-determined system of equations, i.e.,
    the no. of measurements are less than the signal dimension and so the signal
    structure must be exploited through appropriate constraints to obtain a
    unique solution for the problem:
    \begin{align}
    \min_{x \in \Omega} \qquad & L(F({x});{y})\label{eq:CS_prom2}
    \end{align}
    Where $L(F({x});{y})$ is an appropriate loss function to be minimized.
\end{definition}

We want to use the range space of the generative model $q(X;w)$ as the
constraint set $\Omega$ that can super-resolve SAR data as well as potentially
be used to solve a compressed sensing problem for SAR data. Specifically, we
want to find a prior function as an ANN based generative model $G$ with
parameters $w_{G}$, that generates random samples of $X$ (SAR magnitude images)
using a low-dimensional latent rv $Z$, i.e. sampling from $G(Z;w_{G})$ is a good
approximation of sampling from $p(X)$. This implicitly constrains samples $X$ on
a low-dimensional manifold while having the flexibility to adapt the basis to
any dataset, unlike commonly used sparsity priors. Therefore, $G$ can be used as
the learned projection function $\mathcal{P}_{\Omega}$ in a Projected Gradient
Descent optimization to find the solution in the constraint set $\Omega$. We
show that a specially designed $G$ can itself be used for super-resolving SAR
images. The dataset of all samples is split
into 3 subsets $\mathcal{D}_{train}$, $\mathcal{D}_{val}$ and
$\mathcal{D}_{test}$ for model training, cross-validation and evaluation
of $G$ respectively.

\subsection{Related Work}\label{dsp.RW}
A  survey of trained and untrained DL
methods to solve the Inverse problems is given in \cite{ongie2020deep}. Data-driven models can be broadly
classified as end-to-end models that are agnostic to the measurement operator and deep generative priors
that model the distribution representing the constraint set estimated
from a corpus of data. We'll discuss the approaches based on the latter briefly.

Theoretical guarantees are established on the number of measurements and
conditions on the generator network for successful reconstruction
in~\cite{liu2020information}. A generative model is estimated on a large corpus
of natural images in~\cite{bora2017compressed}. Given the generator, which is
implemented as a generative adversarial network or variational auto-encoder, and
the measurements we can infer the image by solving~\eqref{eq:CS_prom2}. The
projection operator $\mathcal{P}_{\Omega}$ for the set of natural images
$\Omega$ is learned from a large corpus of data using an adversarial network
in~\cite{oneNetSolveSankaranarayanan_2017}. This projection operator is used in
solving any inverse problem in a plug and play manner since the projection
operator has enough capacity to model the complicated non-convex set of natural
images. Theoretical guarantees for the generative prior are established
in~\cite{hand2019global}. It is shown that if the entries of the generator
function is near-random and the number of weights in each layer increases with
the width then the loss function landscape contains descent directions to the
global optima. The deep-geometric prior is extended to MR imaging
in~\cite{kelkar2020compressible}. Invertible generators are used. Since the
dimensionality of the latent code is large and equivalent to the signal
dimension, a block-wise structure is imposed on the structure of the latent
variable. The idea of generative prior is extended to video-sequences in a novel
way to synthesize video from a sub-sample of image frames
in~\cite{hyder2020generative}. The image sequences in time are denoted by
$X=\mathbf{x}_1,\mathbf{x}_2,\cdots,\mathbf{x}_T$. It is shown that if all the
images are obtained from the same generator than the network parameters $\theta$
can be fixed and for each image in the sequence a latent code can be estimated
to succinctly represent the video. The latent code sequence
$Z=[\mathbf{z}_1,\mathbf{z}_2,\cdots,\mathbf{z}_T]$ compactly represents the
image. The smoothness in the image sequences of a video can be imposed by
imposing a smoothness constraint or a low-rank constraint on the latent code $Z$
while jointly estimating in the prediction step. Furthermore if images are
dissimilar then the network parameters $\theta$ can also be re-trained to
capture the variability. It is shown that missing frames can be synthesized
using interpolation in the latent space. Next, we consider the regime when no or
limited training data is available, where the generator is estimated for each
image.

Deep image prior~\cite{deepImgePrior2017} presents a non-trained version of the
generator discussed in previous section. The latent code is assumed fixed and
can be chosen arbitrarily but the network parameters are optimized to represent
the image. It is hypothesized that the structure of the network imposes a strong
regularizer or prior on the image. This strong prior is shown to have high
impedance to noise and uncorrelated samples. Therefore, the optimization problem
solved is 
\begin{align}
\min_{\theta} L( g(\mathbf{z},\theta) ,\mathbf{y},F ).
\end{align}
The main problem is over-fitting it is shown that early-stopping is necessary to
capture the de-noised image. This number of iterations serves as a
hyper-parameter that needs to be carefully chosen. Different neural network
architectures are explored and it is empirically shown that as the network is
over-parameterized the performance improves because the capacity of network is
flexible to learn the image. 

The works in~\cite{tirer2020back}, and~\cite{zukerman2020bp} propose an
alternative loss function termed as the backprojection loss and demonstrate the
efficacy theoretically as well as empirically for multiple imaging  linear
inverse problems such as de-blurring and super-resolution as a function of the
condition-number of the measurement operator. The loss function considered is
\begin{align}
    L(\theta) = \lVert F^{+} (\mathbf{y}- F g(\mathbf{x},\theta) )\rVert^2,
\end{align}
where $ F^{+}$ denotes the pseudo-inverse of the linear measurement operator. 
Theoretical guarantees for compressive sensing are established for deep priors
for~\cite{heckel2020compressive}, and~\cite{heckel2019regularizing}. It is shown
that the number of measurements required are similar to the traditional
requirements established in compressive sensing. Empirically, these methods are
shown to perform better than $\ell_1$ and $TV$ norm regularization on fastMRI
dataset. 
\section{Method}\label{dsp.method}
Building upon eq. \ref{eq:CS_prom2}, squared Euclidean $L_2$ norm is commonly
used for the loss function, i.e., $L(F({x});{y})=\lVert {y} -
F({x})\rVert^2_2$ under the assumption that $\eta$ is additive Gaussian noise.
Sparsity in some known basis $\Phi$, such as Fourier or Wavelet basis, is the
most widely used constraint on $x$,
achieved by adding $L_1$ regularization to sparse vector $\tilde{x}$
where $\tilde{x}$ is such that $x=\Phi\tilde{x}$. This derivation involves a convex
relaxation as well as $\Phi$ must follow the restricted isometry property and
\cite{baraniuk2010modelbased} can be referred for details.
Instead of this sparsity assumption, we follow a different approach. The problem
in eq. \ref{eq:CS_prom2} can also be solved by finding a convex
relaxation of the constraint set $\Omega$ and using a projected gradient method as
\begin{align}
    \hat{x}_{mp}&=\argmin_x \lVert {y} - F({x})\rVert^2_2 \nn \\
    &=\argmin_{m,p} \lVert {y} - F({m\odot\exp(jp)})\rVert^2_2\label{eq:mp} \\
    \hat{x}_\Omega&=\mathcal{P}_{\Omega}(\hat{x}_{mp})=\argmin_{x\in \Omega} \lVert \hat{x}_{mp} - x\rVert^2_2\label{eq:omega}
\end{align}

Where $m,p$ are magnitude and phase of complex-valued $x$ respectively alongwith
subscripts (if any), $\mathcal{P}_{\Omega}$ denotes the projection operator onto
set $\Omega$ Alternating between these two steps will find the desired solution
if $\Omega$ and $F$ were convex starting from an appropriate initial condition
$x_0$. However, like most $x$ of interest, we don't know $\Omega$ for SAR data.
Therefore, similar to Projected Gradient Descent GAN by Shah and Hegde
\cite{shah2018solving}, we propose to learn $\Omega$ as the range-space of an
ANN based Generative model $G(Z;w_{G})$. Here, $Z$ as much lower dimensionality
than $x$. Moreover, we only constrain the magnitude $m$ of $x$ using
$G(Z;w_{G})$ and allow phase $p$ to be unconstrained without projection.
Assuming that we have obtained such a $G(Z;w_{G})$ that well approximates the
probability distribution $P(m)$, projection $\hat{x}_\Omega$ becomes
\begin{align} \nonumber
    \hat{x}_\Omega&=\mathcal{P}_{\Omega}(\hat{m}_{mp}\odot \exp(j\hat{p}_{mp}))&=\mathcal{P}_{G}(\hat{m}_{mp})\odot \exp(j\hat{p}_{mp})\\
    &=\hat{m}_G\odot \exp(j\hat{p}_{mp})\label{eq:x_proj}
\end{align}

Where $\hat{m}_G$ is the projection of magnitude $\hat{m}_{mp}$ derived using
$G$. Here, the magnitude $m$ of the structured signal $x$ is assumed to be
sampled from a low-dimensional manifold with a latent variable $z$. To find $\hat{m}_G$, we first optimize over $z$ starting from
initial condition $z_0=0$ as 
\begin{align}
    z^*_z&=\argmin_{z} \lVert \hat{m}_{mp} - G(z;w_{G})\rVert^2_2\label{eq:z}
\end{align}

However, since $G$ is a non-linear ANN, range-space of $G(Z;w_{G})$ is
non-convex. Therefore, solving optimization problem in eq. \ref{eq:z} only
guarantees local-minima. Hence, it will be sensitive to initialization and may
yield inconsistent projections. Motivated from works of Bojanowski et al.
\cite{bojanowski2019optimizing} and Wu et al. \cite{wu2019deep}, we attempt to
alleviate this problem by making $G(Z;w_{G})$ aware of this inversion task
during its training phase. This is described in section \ref{dsp.model}. 

Additionally, we borrow important ideas from similar work of IAGAN
\cite{hussein2019imageadaptive}. Even after significant progress in Deep
Generative Modeling, such models are still an approximation of the true
distribution due to limited representation capabilities of ANN. Therefore,
magnitudes $m$ of many
samples $x$ may not even belong to the range-space of $G(Z;w_{G})$. To mitigate
this, IAGAN proposes to do image-adaptive projections, i.e. optimizing over both
latent vector $z$ and ANN weights $w$ when projecting. This adds the following step to eq.
\ref{eq:z} as
\begin{align}
    (z^*_{zw},w^*_{G})&=\argmin_{z,w} \lVert \hat{m}_{mp} - G(z;w)\rVert^2_2\label{eq:zw}
\end{align}
Where $z$, $w$ are initialized as $z_0=z^*_z$, $w_0=w_G$ respectively while optimizing eq. \ref{eq:zw}.
Finally, the magnitude projection required for eq. \ref{eq:x_proj} is $\hat{m}_{G}=\mathcal{P}_{G}(\hat{m}_{mp})=G(z^*_{zw};w^*_{G})$.
We call our complete algorithm as Image Adaptive Projected Gradient Descent 
WGAN and it is mentioned in algorithm \ref{alg:IAPGD}.


\begin{algorithm}
    \caption{IAPGD WGAN}\label{alg:IAPGD}
    \begin{algorithmic}[1]
    \Require $n,n_{mp},n_{z},n_{zw} \geq 0, y, F, G(z;w_{G}),x_0$
    \While{$n \neq 0$}
        \State $\hat{x}_{mp} \gets \sqb{\argmin_{m,p} \lVert {y} - F({m\odot\exp(jp)})\rVert^2_2}_{n_{mp}}$
        \State $\hat{m}_{mp} \gets \abs{\hat{x}_{mp}}$
        \State $z_0 \gets 0$\label{alg:IAPGD:z_init}
        \State $z^*_z \gets \sqb{\argmin_{z} \lVert \hat{m}_{mp} - G(z;w_{G})\rVert^2_2}_{n_z}$
        \State $(z_0,w_0) \gets (z^*_z,w_G)$
        \State $(z^*_{zw},w^*_{G}) \gets \sqb{\argmin_{z,w} \lVert \hat{m}_{mp} - G(z;w)\rVert^2_2}_{n_{zw}}$
        \State $\hat{m}_{G} \gets G(z^*_{zw};w^*_{G})$\label{alg:IAPGD:m_hat}
        \State $\hat{x}_{\Omega} \gets \hat{m}_G\odot \exp(j\hat{p}_{mp})$
        \State $x_0 \gets \hat{x}_{\Omega}$
        \State $n \gets n - 1$
    \EndWhile
    \end{algorithmic}
\end{algorithm}

Where the notation $\sqb{\argmin_x \mathcal{L}(x)}_{n_{x}}$ refers to minimizing the
loss $\mathcal{L}(x)$ for $n_{x}$ steps iteratively using a variant of
Stochastic Gradient Descent (SGD).

We propose a special hierarchical architecture of a WGAN that jointly generates
magnitude of multiple resolutions of the same SAR image. This idea of jointly
modeling multiple resolutions was inspired from the Progressive GAN by Karras et
al. \cite{karras2018progressive} though their motivations were different. We aim
to exploit this hierarchical structure for super-resolution i.e. to find a
higher resolution image given its lower resolutions. Suppose we are given a
dataset containing 4 exponentially increasing resolution images' magnitude
$m^{r_1},m^{r_2},m^{r_3},m^{r_4}$ where resolution of $m^{r_{i+1}}$ is twice of
$m^{r_i}$. Then our WGAN $G(Z;w_G)$ models the joint probability distribution
$P(m^{r_1},m^{r_2},m^{r_3},m^{r_4})$. Now if we are given a new sample from
$P(m^{r_1},m^{r_2},m^{r_3})$ i.e. of the 3 lower resolutions, we can use
$G(Z;w_G)$ and steps \ref{alg:IAPGD:z_init} to \ref{alg:IAPGD:m_hat} of
algorithm \ref{alg:IAPGD} to project it onto the joint data manifold of
$P(m^{r_1},m^{r_2},m^{r_3},m^{r_4})$ by finding a common $(z^*_{zw},w^*_{G})$
pair. The highest resolution $r_4$ image is then obtained by a simple forward
pass $\hat{m}^{r_4}_G=G(z^*_{zw},w^*_{G})$. The overall algorithm is therefore 
Projecting from a Multi-Resolution SAR Prior (or MrSARP) and is summarized in algorithm
\ref{alg:ProMr}. Note that $G(z;w_{G})^{r_i}$ denotes ${r_i}$ resolution output
from $G(z;w_{G})$ and $G(z;w_{G})^{r_4}$ is simply dropped in step
\ref{alg:ProMr:Opz}.

\begin{algorithm}
    \caption{Projecting from MrSARP}\label{alg:ProMr}
    \begin{algorithmic}[1]
    \Require $n_{z},n_{zw} \geq 0, m^{r_1},m^{r_2},m^{r_3}, G(z;w_{G})$
    \State $z_0 \gets 0$
    \State $z^*_z \gets \sqb{\argmin_{z} \Sigma_{i=1}^3\lVert m^{r_i} - G(z;w_{G})^{r_i}\rVert^2_2}_{n_z}$\label{alg:ProMr:Opz}
    \State $(z_0,w_0) \gets (z^*_z,w_G)$
    \State $(z^*_{zw},w^*_{G}) \gets \sqb{\argmin_{z,w} \Sigma_{i=1}^3\lVert m^{r_i} - G(z;w)^{r_i}\rVert^2_2}_{n_{zw}}$
    \State $\hat{m}^{r_4}_G \gets G(z^*_{zw};w^*_{G})^{r_4}$
    \end{algorithmic}
\end{algorithm}

\section{Model}\label{dsp.model}

\begin{figure*}
    \centering
    \subcaptionbox{Generator\label{fig:dsp.model.gen}}
    [.49\linewidth]{\includegraphics[width=0.37\textwidth,keepaspectratio]{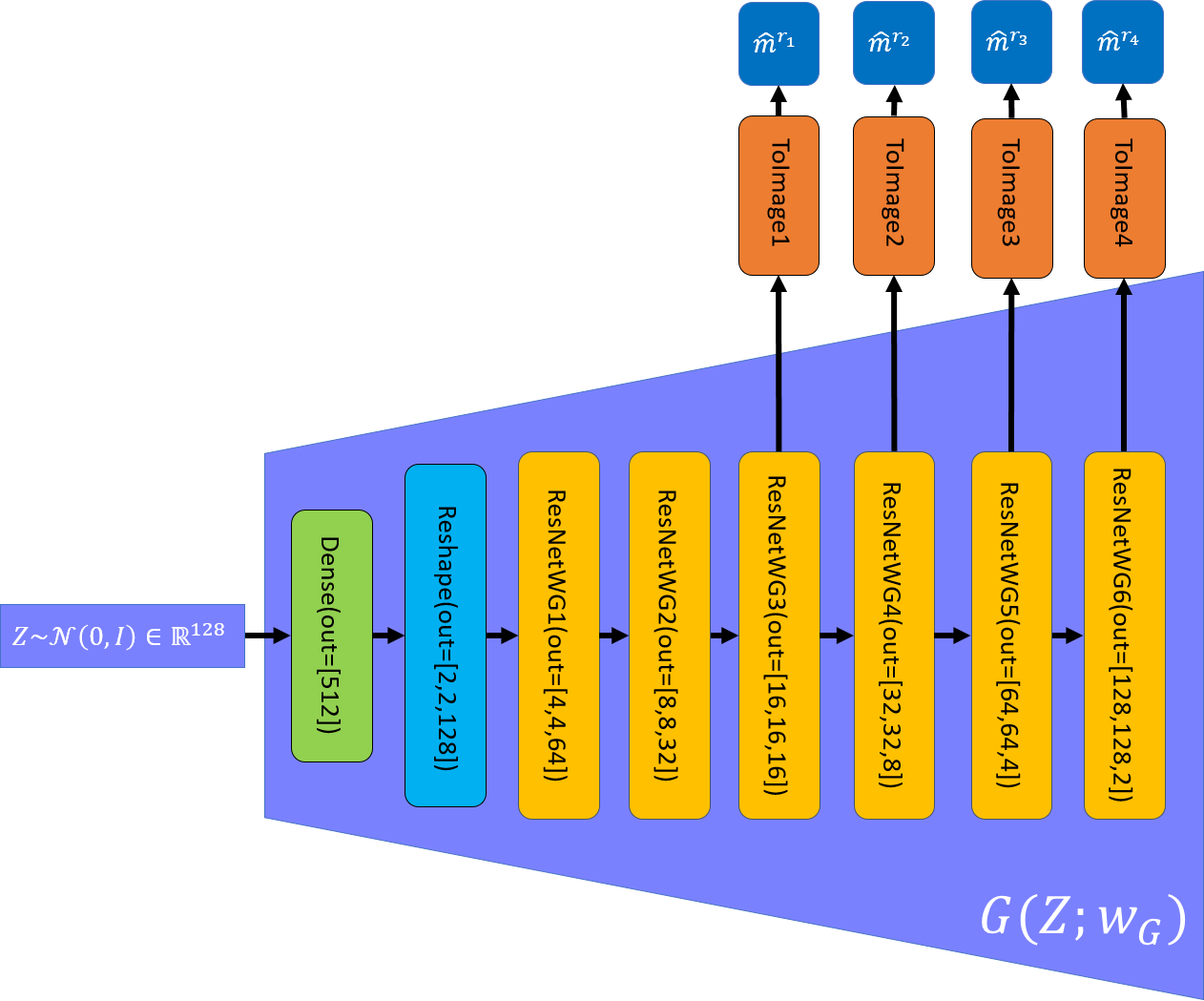}}%
    \subcaptionbox{Critic\label{fig:dsp.model.disc}}
    [.42\linewidth]{\includegraphics[width=0.3\textwidth,keepaspectratio]{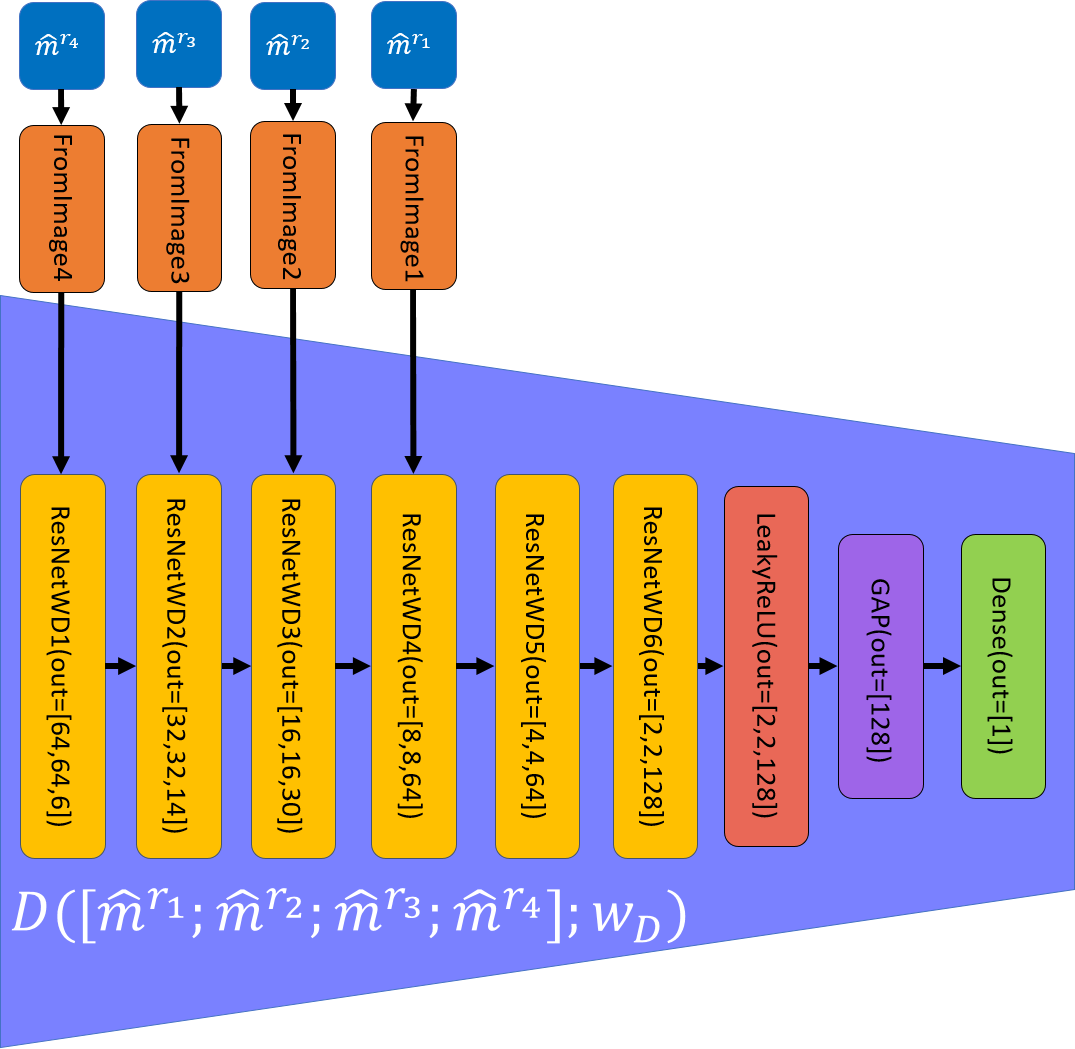}}
    \caption{Hierarchical Model Architecture of MrSARP }\label{fig:dsp.model}
\end{figure*}

MrSARP consists of a WGAN-GP \cite{gulrajani2017improved} with a Hierarchical
architecture. Our architecture is inspired from the CIFAR-10 ResNet architecture used by Gulrajani et
al. \cite{gulrajani2017improved} and ProGAN \cite{karras2018progressive}. We use
the ResNet block from the former and the idea of FromImage/ToImage layers at
various ResNet input/output features from latter. The schematic of the
hierarchical model is shown in figure \ref{fig:dsp.model}. The output shapes of
layers are specified. The ResNet blocks in the 
generator use Batch-Normalization layers and have nearest-neighbour upsampling after the input. The ResNet blocks in the 
critic uses Layer-Normalization layers and have an average pooling based
downsampling before the final output, and channel-wise concatenation in case of
multiple inputs. The FromImage layer is a 2D convolution layer with 1 channel
input (image) and 2 channel output (features). The
ToImage layer comprises slicing first 2 channels of input features, Batch-Normalization, ReLU activation and 2D convolution
layer with 1 channel output (image) after tanh activation. All convolutional
layers in the generator have kernel-size as 3 and ReLU activation function. All convolutional
layers in the critic have kernel-size as 3 and LeakyReLU activation function
with 0.2 slope. The dense (or fully-connected) layers in both, generator and
critic have linear activation and the GAP referes to Global Average Pooling
operation. The generator $G(Z;w_G)$ of MrSARP generates samples from the
unconditional joint
distribution of 4 resolutions of SAR images $P(m^{r_1},m^{r_2},m^{r_3},m^{r_4})$
where resolution of $m_{r_i}$ is $2^{i+3}\times 2^{i+3}$ and bandwidth is
$125.2^{i} Mhz.$.

WGAN-GP was chosen because of its advantages over traditional GANs, especially
meaningful loss curves for cross-validation and reliable training as
demonstrated by Gulrajani et al. \cite{gulrajani2017improved}. WGAN-GP, like all
GANs, require an extra critic ANN $D(\cdot,\cdot;w_{D})$ to be learned
simultaneously to aid the learning of the generator. Both, the critic and
generator are trained alternately using gradient descent steps. Latent variable $Z \sim
\mathcal{N}(0,I)$ as in a conventional GAN.

Since MrSARP would primarily be used for inverting and projecting on
data-manifold instead of sampling, we think it is essential to inform the
generator $G$ about this task during training. Inspired from Wu et al's work
\cite{wu2019deep}, we add Model Agnostic Meta Learning (MAML) based regularizing
loss term to the overall loss functions used for training $G$. Proposed by Finn
et al. \cite{finn2017modelagnostic}, MAML is a general-purpose meta-learning method to
adapt parameters $w$ of a statistical model to a number of tasks as long as the
loss function $\mathcal{L}(\mathcal{T};w)$ for the task $\mathcal{T}$ is differentiable.
Since our primary task is inversion using step \ref{alg:ProMr:Opz} of algorithm
\ref{alg:ProMr}, we perform a small number $n_z=5$ iterations of this
optimization step to get $z^*$ and use
residual error on all 4 channels as our MAML loss $\mathcal{L}_{MAML}$. Since
the $Z \sim \mathcal{N}(0,I)$ with high dimensionality $d_Z$, the samples
lie near the hypersphere of $\sqrt{d_Z}$. To enforce such a constraint on $z^*$, we
additionally project $z^*$ on to the $\sqrt{d_Z}$ hypersphere after every
SGD step using projection operator $\mathcal{P_S}$.  The complete loss
functions are as follows.


\begin{align}
	\mathcal{L} &= 
	\begin{cases}
		-\mathcal{L}_{W} + \lambda_1\mathcal{L}_{GP}&\text{for } D \\
		\mathcal{L}_{W}  + \lambda_2\mathcal{L}_{MAML}&\text{for } G
	\end{cases}\label{eq:dsp.loss}
\end{align}

\begin{align}
	\mathcal{L}_{W} &=  \E_{m}\sqb{D(m);w_D}-\E_{Z}\sqb{D(G(Z;w_{G});w_D)} \nn
	\\
	\mathcal{L}_{GP} &= \E_{\bar{m}}(\norm{\nabla_{\bar{m}}D(\bar{m};w_D)}_2-1)^2 \nn
	\\
    \mathcal{P_S}(z^*)&=\frac{z^*}{\min(\norm{z^*}_2,\sqrt{d_Z})}.\sqrt{d_Z} 
    \\
    z^*_S &= \sqb{\mathcal{P_S}\pb{\argmin_{z} \Sigma_{i=1}^3\lVert m^{r_i} - G(z;w_{G})^{r_i}\rVert^2_2}}\label{z_star}
	\\
    \mathcal{L}_{MAML} &= \E_{m}\sqb{\Sigma_{i=1}^4 \lVert m^{r_i} - G(z^*_S;w_{G})^{r_i}\rVert^2_2}
\end{align}

Where $m \sim P(m^{r_1},m^{r_2},m^{r_3},m^{r_4})$, $\bar{m} \sim
r(\bar{M})$ is uniform-sampling along straight lines between pairs of sampled
points $m$, $G(z;w_{G})$. All expectations are approximated
using corresponding empirical means. Motivated by \cite{gulrajani2017improved},
we employ negative critic loss as our primary metric for model selection. Hence,
the weights of the final model are set to their values corresponding to the
epoch where the smallest negative critic loss on $\mathcal{D}_{val}$ was
achieved. For the hyperparameters, we use $\lambda_1=10$ (as in
\cite{gulrajani2017improved}) and set $\lambda_2=1000$ to
approximately balance the two regularizing loss term $L_{MAML}$ 
with the WGAN loss term $\mathcal{L}_{W}$ in magnitude.

\section{Experimental Setup}\label{dsp.exp}

We use the Tensorflow (2.1) \cite{abadi2016tensorflow} deep-learning Python
library for our ANN implementations. Unless mentioned otherwise, the Adam
optimizer from Tensorflow with default parameters is used for most
optimizations. We use NVIDIA GeForce RTX 2080 Ti GPU alongwith Intel Xeon CPU as
our primary computation hardware.

\subsection{CVDomes Dataset}\label{dsp.data}
We use the phase histories obtained from simulation of back-scattered energy
from civilian vehicles in \cite{dungan2010civilian}. We consider only the HH
polarization measurement in our experiment.  The image at a bandwidth B and
azimuth span $\Delta \theta$ is obtained by backprojection method using a
Hamming window to suppress the side-lobes. We  spotlight on a square patch of
$9m\times9m$.  To generate 4 different resolutions $r{1:4}$, we repeat this
process for 4 bandwidths $125.2^i$ and corresponding pixel resolutions
$2^{i+3}\times2^{i+3}$ for $i\in {1,2,3,4}$.

\subsection{Pre-Processing}\label{dsp.data.prepro}
We work with magnitude images only for MrSARP. Hence, we first find the absolute
values of complex-valued SAR images and then perform min-max normalization of
every image individually to restrict their values in range $[-1,1]$. We then
upsample all lower resolution images to the highest resolution of $128\times128$
with nearest-neighbor upsampling and combine all 4 resolutions into a single 4
channel image. The lower resolution images are appropriately downsampled using
average pooling before their input into the critic. This is only done to
simplify multi-resolution architecture implementation and is based on the simple
fact that nearest-neighbor upsampling followed by average-pooling gives identity
function.

\section{Results}\label{dsp.res}
We present some results of this study in this section. After training
generator $G$ as described in section \ref{dsp.model} above, we
use algorithm \ref{alg:ProMr} to find the super-resolved images
$\hat{m}_G^{r_4}$ for the unseen lower resolution samples in
$\mathcal{D}_{test}$. We compare these images with the available ground truth
images $m^{r_4}$ qualitatively as well as using three quantitative metrics viz. PSNR (Peak Signal-to-Noise Ratio), NMSE (Normalised Mean
Squared Error) as defined in \cite{karimi2018nonparametric} and SSIM (Structural
Similarity Index) proposed in \cite{wang2004image}. The use of these metrics is
motivated from the evaluation schemes used in existing literature on
super-resolution of SAR images e.g. \cite{he2012learning}. These metric used are defined
as follows.

\begin{align}
    MSE&=\frac{1}{N_m}\norm{m^{r_4}-\hat{m}^{r_4}}_F^2 \nn \\
    PSNR&=10\log_{10}\pb{\frac{(\max(m^{r_4})-\min(m^{r_4}))^2}{MSE}} \\
    NMSE&=\frac{\norm{m^{r_4}-\hat{m}^{r_4}}_F^2}{\norm{m^{r_4}}_F^2} \\
    SSIM&=l(x,y).c(x,y).s(x,y)\\
    l(x,y)&=\pb{\frac{2\mu_x\mu_y+C_1}{\mu_x^2+\mu_y^2+C_1}}\nn \\ c(x,y)&=\pb{\frac{2\sigma_x\sigma_y+C_2}{\sigma_x^2+\sigma_y^2+C_2}}\nn \\
    s(x,y)&=\pb{\frac{\sigma_{xy}+C_3}{\sigma_x\sigma_y+C_3}}\nn
\end{align}

where $N_m$ are total number of pixels in image $m^{r_4}$, $\mu_x,\mu_y$ are
empirical means of patches $x,y$ respectively, $\sigma_x,\sigma_y$ are sample
standard deviations of patches $x,y$ respectively, $\sigma_{xy}$ is sample
cross-correlation of $x,y$ after mean subtraction and $C_{1:3}$ are small constants
added for numerical stability. SSIM is calculated on smaller local
patches $x,y$ of size $7\times7$ and the mean SSIM is calculated for every image
comparison pair $m^{r_4},\hat{m}^{r_4}$.

We also compare $r_4$ resolution estimate $\hat{m}_N^{r_4}$ obtained from
nearest neighbor upsampling of $m^{r_3}$ and estimates $\hat{m}_L^{r_4}$
obtained from using the popular LASSO method. The LASSO method is used to obtain
the sparse scattering center representation of the vehicle at the resolution
${r_3}$. The sparse representation is projected back into the phase history
measurement domain using the SAR forward operator. These measurements are
converted to SAR imagery using the backprojection method with the hamming window
at resolution ${r_4}$. All the evaluations are done on 504 images from the
unseen $\mathcal{D}_{test}$.

\begin{figure*}
    \centering

    \subcaptionbox{Image 1\label{fig:dsp.res.qual3}}
    [.72\linewidth]{\includegraphics[width=0.71\textwidth,keepaspectratio]{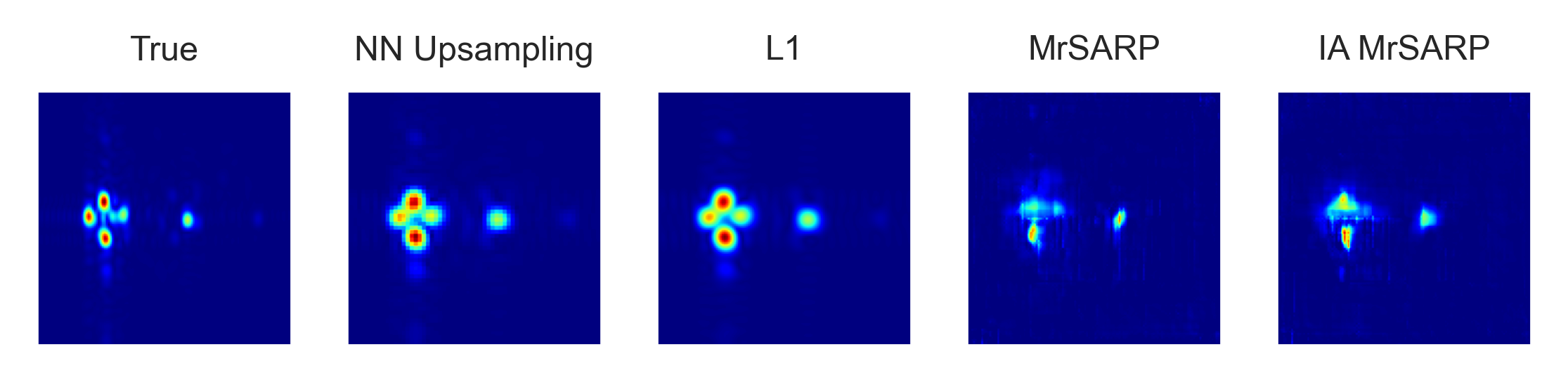}}

    \subcaptionbox{Image 2\label{fig:dsp.res.qual4}}
    [.72\linewidth]{\includegraphics[width=0.71\textwidth,keepaspectratio]{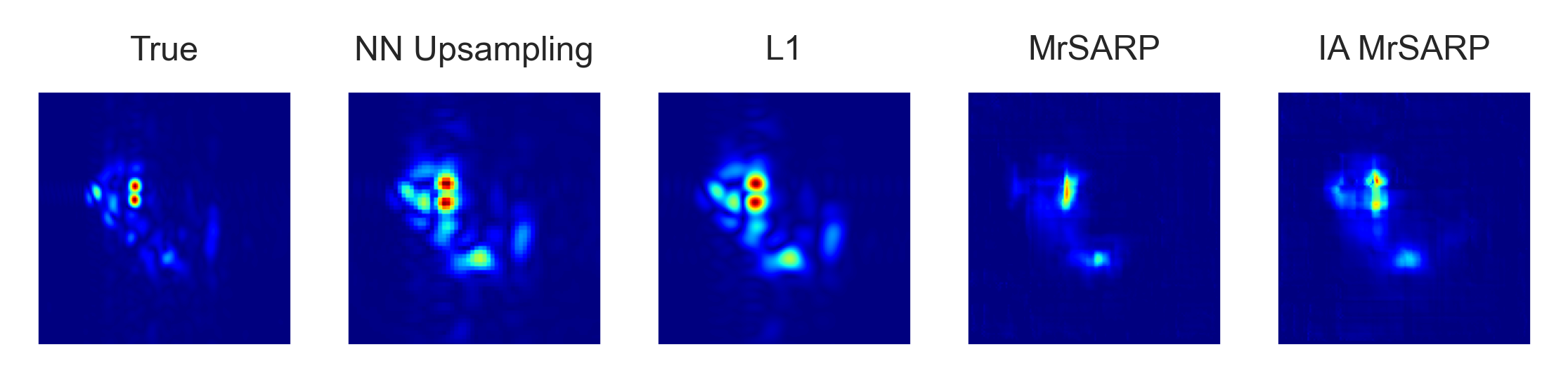}}
    \subcaptionbox{Image 3\label{fig:dsp.res.qual5}}
    [.72\linewidth]{\includegraphics[width=0.71\textwidth,keepaspectratio]{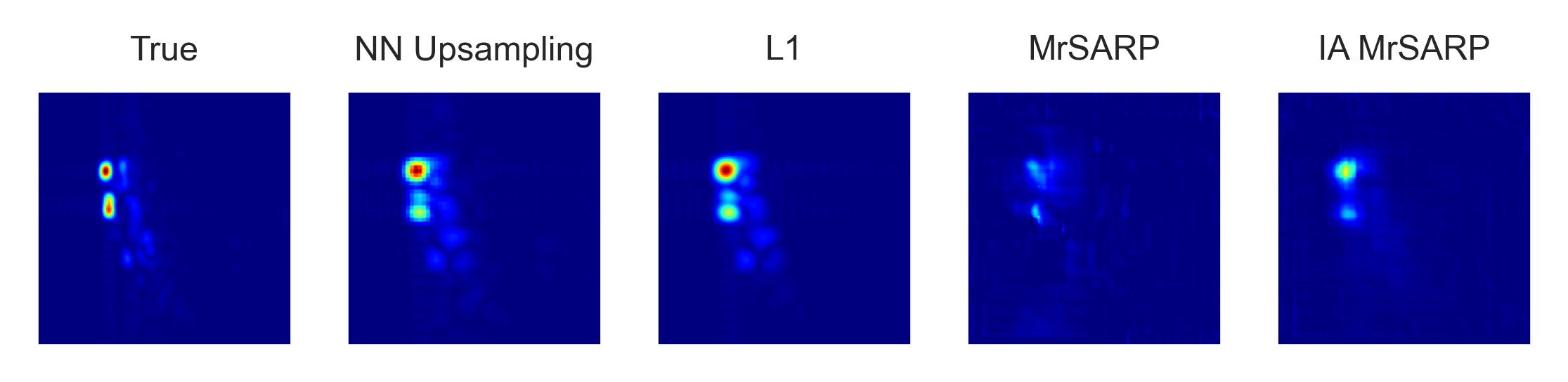}}

    \caption{Qualitative comparison of samples super-resolved using different methods}\label{fig:dsp.res.qual}
\end{figure*}

Figure \ref{fig:dsp.res.qual} shows some samples from $\mathcal{D}_{test}$ that
are super-resolved using different methods. There is a qualitative similarity
between the Nearest Neighbor upsampled and $L_1$ recovered images. Both methods
differ significantly from resulting images of MrSARP.

 These quantitative results are presented in table
\ref{table:dsp.res}.  Addition of Image-Adaptive steps to MrSARP improves
performance quantitatively for all 3 metrics. In fact, among all the methods
tested, Image-Adaptive MrSARP results
perform best in terms of both, $NMSE$ and $PSNR$. However, they are inferior to
both Nearest Neighbor Upsampling and $L_1$ recovery in terms of SSIM indicating potential bias in amplitude for MrSARP. 
\begin{table}
	\begin{center}
	\caption{Quantitative evaluation of Super-resolution performance.\label{table:dsp.res}}
	\begin{tabular}{ | c | c | c | c | }
    \hline
      Method & $NMSE \downarrow$ & $PSNR \uparrow$ & $SSIM \uparrow$ \\\hline
      Nearest-Neighbor Upsampling & 9.071 & 27.27 & 0.931 \\\hline
      $L_1$ recovery & 8.698 & 27.638 & \textbf{0.937} \\\hline
      MrSARP & 6.92 & 29.785 & 0.9 \\\hline
      Image-Adaptive MrSARP & \textbf{6.003} & \textbf{30.963} & 0.918 \\\hline
    \end{tabular}
	\end{center}
\end{table}

\section{Conclusion}\label{dsp.conc}
In this paper, we showed how a GAN with hierarchical architecture can
jointly model the distribution of multiple resolutions of magnitude SAR images.
We further showed how this GAN, called MrSARP, be used to for super-resolving SAR
images. We saw some improvements over baselines of nearest neighbor upsampling as
well as $L_1$ recovery in terms of $NMSE$ and $PSNR$ values 
but there is significant scope of further improvements in terms of perceptual
quality. This was indicated by MrSARP's inferior performance in terms of SSIM. 
Furthermore, we plan to utilize this generative model and the proposed algorithm
\ref{alg:IAPGD} to regularize inverse problems in SAR imaging with structured
interrupted measurements. 

\section*{Acknowledgements}
This research was partially supported by NSF grants CNS-1823070
CBET-2037398 and NIH Grant P41EB028242
\bibliographystyle{IEEEtran}
\bibliography{references}

\begin{thebibliography}{10}
\providecommand{\url}[1]{#1}
\csname url@samestyle\endcsname
\providecommand{\newblock}{\relax}
\providecommand{\bibinfo}[2]{#2}
\providecommand{\BIBentrySTDinterwordspacing}{\spaceskip=0pt\relax}
\providecommand{\BIBentryALTinterwordstretchfactor}{4}
\providecommand{\BIBentryALTinterwordspacing}{\spaceskip=\fontdimen2\font plus
\BIBentryALTinterwordstretchfactor\fontdimen3\font minus
  \fontdimen4\font\relax}
\providecommand{\BIBforeignlanguage}[2]{{%
\expandafter\ifx\csname l@#1\endcsname\relax
\typeout{** WARNING: IEEEtran.bst: No hyphenation pattern has been}%
\typeout{** loaded for the language `#1'. Using the pattern for}%
\typeout{** the default language instead.}%
\else
\language=\csname l@#1\endcsname
\fi
#2}}
\providecommand{\BIBdecl}{\relax}
\BIBdecl

\bibitem{ongie2020deep}
G.~Ongie, A.~Jalal, C.~A. M. R.~G. Baraniuk, A.~G. Dimakis, and R.~Willett,
  ``Deep learning techniques for inverse problems in imaging,'' \emph{IEEE
  Journal on Selected Areas in Information Theory}, 2020.

\bibitem{liu2020information}
Z.~Liu and J.~Scarlett, ``Information-theoretic lower bounds for compressive
  sensing with generative models,'' \emph{IEEE Journal on Selected Areas in
  Information Theory}, 2020.

\bibitem{bora2017compressed}
A.~Bora, A.~Jalal, E.~Price, and A.~G. Dimakis, ``Compressed sensing using
  generative models,'' in \emph{International Conference on Machine Learning},
  2017, pp. 537--546.

\bibitem{oneNetSolveSankaranarayanan_2017}
J.~H.~R. Chang, C.~Li, B.~P{\'o}czos, B.~V.~K. Vijaya~Kumar, and A.~C.
  Sankaranarayanan, ``One network to solve them all \textemdash{} {{Solving}}
  linear inverse problems using deep projection models,'' in \emph{2017
  {{IEEE}} International Conference on Computer Vision ({{ICCV}})}, 2017, pp.
  5889--5898.

\bibitem{hand2019global}
P.~Hand and V.~Voroninski, ``Global guarantees for enforcing deep generative
  priors by empirical risk,'' \emph{IEEE Transactions on Information Theory},
  vol.~66, no.~1, pp. 401--418, 2019.

\bibitem{kelkar2020compressible}
V.~A. Kelkar, S.~Bhadra, and M.~A. Anastasio, ``Compressible latent-space
  invertible networks for generative model-constrained image reconstruction,''
  \emph{arXiv preprint arXiv:2007.02462}, 2020.

\bibitem{hyder2020generative}
R.~Hyder and M.~S. Asif, ``Generative models for low-dimensional video
  representation and reconstruction,'' \emph{IEEE Transactions on Signal
  Processing}, vol.~68, pp. 1688--1701, 2020.

\bibitem{deepImgePrior2017}
V.~Lempitsky, A.~Vedaldi, and D.~Ulyanov, ``Deep image prior,'' in \emph{2018
  {{IEEE}}/{{CVF}} Conference on Computer Vision and Pattern Recognition},
  2018, pp. 9446--9454.

\bibitem{tirer2020back}
T.~Tirer and R.~Giryes, ``Back-projection based fidelity term for ill-posed
  linear inverse problems,'' \emph{IEEE Transactions on Image Processing},
  vol.~29, pp. 6164--6179, 2020.

\bibitem{zukerman2020bp}
J.~Zukerman, T.~Tirer, and R.~Giryes, ``{{BP-DIP}}: {{A}} backprojection based
  deep image prior,'' \emph{arXiv preprint arXiv:2003.05417}, 2020.

\bibitem{heckel2020compressive}
R.~Heckel and M.~Soltanolkotabi, ``Compressive sensing with un-trained neural
  networks: {{Gradient}} descent finds the smoothest approximation,''
  \emph{arXiv preprint arXiv:2005.03991}, 2020.

\bibitem{heckel2019regularizing}
R.~Heckel, ``Regularizing linear inverse problems with convolutional neural
  networks,'' \emph{arXiv preprint arXiv:1907.03100}, 2019.

\bibitem{baraniuk2010modelbased}
\BIBentryALTinterwordspacing
R.~G. Baraniuk, V.~Cevher, M.~F. Duarte, and C.~Hegde, ``Model-{{Based
  Compressive Sensing}},'' \emph{IEEE Transactions on Information Theory},
  vol.~56, no.~4, pp. 1982--2001, Apr. 2010. [Online]. Available:
  \url{http://arxiv.org/abs/0808.3572}
\BIBentrySTDinterwordspacing

\bibitem{shah2018solving}
V.~Shah and C.~Hegde, ``Solving {{Linear Inverse Problems Using Gan Priors}}:
  {{An Algorithm}} with {{Provable Guarantees}},'' in \emph{2018 {{IEEE
  International Conference}} on {{Acoustics}}, {{Speech}} and {{Signal
  Processing}} ({{ICASSP}})}, Apr. 2018, pp. 4609--4613.

\bibitem{bojanowski2019optimizing}
\BIBentryALTinterwordspacing
P.~Bojanowski, A.~Joulin, D.~{Lopez-Paz}, and A.~Szlam, ``Optimizing the
  {{Latent Space}} of {{Generative Networks}},'' May 2019. [Online]. Available:
  \url{http://arxiv.org/abs/1707.05776}
\BIBentrySTDinterwordspacing

\bibitem{wu2019deep}
\BIBentryALTinterwordspacing
Y.~Wu, M.~Rosca, and T.~Lillicrap, ``Deep {{Compressed Sensing}},'' May 2019.
  [Online]. Available: \url{http://arxiv.org/abs/1905.06723}
\BIBentrySTDinterwordspacing

\bibitem{hussein2019imageadaptive}
\BIBentryALTinterwordspacing
S.~A. Hussein, T.~Tirer, and R.~Giryes, ``Image-{{Adaptive GAN}} based
  {{Reconstruction}},'' Nov. 2019. [Online]. Available:
  \url{http://arxiv.org/abs/1906.05284}
\BIBentrySTDinterwordspacing

\bibitem{karras2018progressive}
\BIBentryALTinterwordspacing
T.~Karras, T.~Aila, S.~Laine, and J.~Lehtinen, ``Progressive {{Growing}} of
  {{GANs}} for {{Improved Quality}}, {{Stability}}, and {{Variation}},'' Feb.
  2018. [Online]. Available: \url{http://arxiv.org/abs/1710.10196}
\BIBentrySTDinterwordspacing

\bibitem{gulrajani2017improved}
\BIBentryALTinterwordspacing
I.~Gulrajani, F.~Ahmed, M.~Arjovsky, V.~Dumoulin, and A.~Courville, ``Improved
  {{Training}} of {{Wasserstein GANs}},'' \emph{arXiv:1704.00028 [cs, stat]},
  Dec. 2017. [Online]. Available: \url{http://arxiv.org/abs/1704.00028}
\BIBentrySTDinterwordspacing

\bibitem{finn2017modelagnostic}
\BIBentryALTinterwordspacing
C.~Finn, P.~Abbeel, and S.~Levine, ``Model-{{Agnostic Meta-Learning}} for
  {{Fast Adaptation}} of {{Deep Networks}},'' \emph{arXiv:1703.03400 [cs]},
  Jul. 2017. [Online]. Available: \url{http://arxiv.org/abs/1703.03400}
\BIBentrySTDinterwordspacing

\bibitem{abadi2016tensorflow}
\BIBentryALTinterwordspacing
M.~Abadi, P.~Barham, J.~Chen, Z.~Chen, A.~Davis, J.~Dean, M.~Devin,
  S.~Ghemawat, G.~Irving, M.~Isard, M.~Kudlur, J.~Levenberg, R.~Monga,
  S.~Moore, D.~G. Murray, B.~Steiner, P.~Tucker, V.~Vasudevan, P.~Warden,
  M.~Wicke, Y.~Yu, and X.~Zheng, ``\{\vphantom\}{{TensorFlow}}\vphantom\{\}:
  {{A System}} for \{\vphantom\}{{Large-Scale}}\vphantom\{\} {{Machine
  Learning}},'' in \emph{12th {{USENIX Symposium}} on {{Operating Systems
  Design}} and {{Implementation}} ({{OSDI}} 16)}, 2016, pp. 265--283. [Online].
  Available:
  \url{https://www.usenix.org/conference/osdi16/technical-sessions/presentation/abadi}
\BIBentrySTDinterwordspacing

\bibitem{dungan2010civilian}
\BIBentryALTinterwordspacing
K.~E. Dungan, C.~Austin, J.~Nehrbass, and L.~C. Potter, ``Civilian vehicle
  radar data domes,'' in \emph{Algorithms for {{Synthetic Aperture Radar
  Imagery XVII}}}, vol. 7699.\hskip 1em plus 0.5em minus 0.4em\relax {SPIE},
  Apr. 2010, pp. 242--253. [Online]. Available:
  \url{http://www.spiedigitallibrary.org/conference-proceedings-of-spie/7699/76990P/Civilian-vehicle-radar-data-domes/10.1117/12.850151.full}
\BIBentrySTDinterwordspacing

\bibitem{karimi2018nonparametric}
N.~Karimi and M.~R. Taban, ``Nonparametric blind {{SAR}} image super resolution
  based on combination of the compressive sensing and sparse priors,''
  \emph{Journal of Visual Communication and Image Representation}, vol.~55, pp.
  853--865, 2018.

\bibitem{wang2004image}
Z.~Wang, A.~Bovik, H.~Sheikh, and E.~Simoncelli, ``Image quality assessment:
  From error visibility to structural similarity,'' \emph{IEEE Transactions on
  Image Processing}, vol.~13, no.~4, pp. 600--612, Apr. 2004.

\bibitem{he2012learning}
C.~He, L.~Liu, L.~Xu, M.~Liu, and M.~Liao, ``Learning {{Based Compressed
  Sensing}} for {{SAR Image Super-Resolution}},'' \emph{IEEE Journal of
  Selected Topics in Applied Earth Observations and Remote Sensing}, vol.~5,
  no.~4, pp. 1272--1281, Aug. 2012.

\end{thebibliography}



\end{document}